\documentclass[sigconf,nonacm]{acmart}
\usepackage{amsmath}
\usepackage{booktabs}
\usepackage{tabularx}
\setcopyright{none}
\begin{document}
\title[More Features Are Not More Evidence]{More Features Are Not More Evidence: Limits of Training-Free Human Activity Recognition with Jev}
\author{Orhan Konak}
\email{orhan.konak@hpi.de}
\orcid{0000-0003-1884-8029}
\affiliation{\institution{Hasso Plattner Institute, University of Potsdam}\city{Potsdam}\country{Germany}}
\renewcommand{\shortauthors}{Konak}
\begin{abstract}
General-purpose models promise sensor-based decisions without training a task-specific classifier, which could reduce the dependence of Human Activity Recognition (HAR) on labeled data. Yet it remains unclear whether such models can directly interpret deterministic descriptions of physical sensor signals well enough to replace or complement trained HAR models. We study this question using Jev, a fixed general-purpose probabilistic decision model, on 1,800 class-balanced accelerometer windows from WISDM, UCI341, and PAMAP2. Jev receives no labeled examples, retrieval context, or HAR-specific parameter updates. We evaluate three deterministic sensor representations and compare 5,400 Jev decisions with a generative baseline and three supervised HAR models. Jev remains far below supervised recognition, with its strongest representation reaching macro-F1 of 0.038, 0.118, and 0.089 across the three datasets, compared with 0.686 to 0.907 for the supervised models. More numerical features do not improve Jev. Instead, they reduce recognition on all three datasets, while augmenting the same numerical evidence with a deterministic semantic rendering partially recovers performance, although the experiment does not isolate semantics from the accompanying serialization and redundancy changes. Jev is fast and inexpensive to query, but its probabilities are not reliably calibrated for recognition. A post-hoc fusion analysis finds a small improvement on WISDM that does not replicate on UCI341 or PAMAP2. These results show that training-free sensor decisions depend not only on the information available in the signal, but also on whether the model can use the representation through which that information is exposed. The sensor-to-model interface should therefore be treated as part of the model evaluation rather than as a neutral preprocessing step.
\end{abstract}
% \keywords{human activity recognition, wearable sensing, training-free inference, decision models, sensor representation, probabilistic fusion}
\keywords{human activity recognition, wearable sensing, training-free inference, probabilistic decision models, uncertainty estimation, sensor-based machine learning}
\maketitle
\section{Introduction}
\label{sec:introduction}
A cheap, fast activity decision is useful only if it is correct often enough for its intended application. Sensor-based human activity recognition (HAR) conventionally learns the mapping from physical measurements to activity labels using task-specific training data~\cite{lara2013survey}. General-purpose models invite a different workflow. A sensor window can be described to the model together with a set of candidate activities, allowing it to make a decision without training a dedicated sensor classifier. The attraction is clear, especially when measurements are plentiful and annotations are scarce. What remains unclear is how well such an interface actually works for sensor-based activity recognition.

We investigate Jev, the proprietary model introduced by TypeSafe AI under the manufacturer's term \emph{System One Models}~\cite{almeida2026jev}. Jev accepts a state and a bounded set of choices and returns a typed decision and probabilities. We treat this as an observable API contract, not as evidence for a particular internal architecture, reasoning process, or training corpus. The question is whether a fixed general-purpose decision model can use deterministic descriptions of accelerometer signals without HAR-specific training or labeled reference examples.

The results show a clear limitation. Across three datasets, supervised classifiers remain far ahead of Jev and a generative baseline. More numerical features do not help Jev: expanding a 20-feature statistical representation to 32 signal-aware features drives predictions toward a single class. Augmenting the same numerical representation with a deterministic semantic rendering recovers some performance but leaves a large absolute gap. \emph{More sensor information is not necessarily more usable evidence.} Here, that statement describes the tested interfaces; it is not a universal law about feature count or all general-purpose models.

The distinction matters beyond standalone accuracy. Jev's low observed API cost and subsecond median response time make repeated queries feasible, while its poor recognition and imperfect probability validity limit what those queries can support. In a separate exploratory analysis, a small Jev probability contribution can improve a Random Forest (RF) on WISDM, but this pattern does not replicate on the other two datasets. We report the entire fixed fusion curve rather than presenting a retrospectively selected weight as a deployable model.

This paper makes three contributions:
\begin{itemize}
    \item We isolate a strict training-free HAR setting in which a fixed general-purpose decision model receives only a deterministic representation of the current sensor window, a fixed task instruction, and the candidate activity set, without HAR-specific training, labeled reference examples, or retrieval. We evaluate this setting on the same subject-independent benchmark windows as three supervised HAR references and a generative baseline.
    \item We show that information available in a sensor representation is not necessarily usable evidence for a fixed general-purpose decision model. Adding numerical signal descriptors reduces Jev's recognition on all three datasets, whereas augmenting the complete numerical representation with a deterministic semantic rendering partially recovers performance. This identifies the sensor-to-model interface as part of the model evaluation rather than a neutral preprocessing choice.
    \item We characterize the limits and possible complementary value of this interface through probability validation, calibration and selective-prediction analyses, operational measurements, and a separately frozen post-hoc fusion study. Jev remains far below trained HAR models, while the small fusion gain observed on WISDM does not replicate on UCI341 or PAMAP2.
\end{itemize}

% This paper makes three contributions. First, we provide a matched, subject-independent comparison of Jev, a generative baseline, and three supervised HAR models across three accelerometer benchmarks. Second, we show that richer numerical descriptions can reduce recognition, while deterministic semantic augmentation of the same numerical evidence can partially recover it. This result highlights the sensor interface as an experimental factor in its own right. Third, we evaluate the operational and probabilistic behavior of Jev and examine, in a separate post-hoc analysis, whether its probability estimates provide complementary information to a trained HAR classifier.

% This paper contributes:
% \begin{itemize}
% \item A matched, subject-independent comparison of Jev, a generative baseline, and RF, SVM, and 1D-CNN on three accelerometer benchmarks, with frozen inputs and subject-cluster confidence intervals.
% \item An interface-level result: richer numerical descriptors can reduce recognition, whereas deterministic semantic augmentation of the same numerical evidence can partially restore it. We document the executed R3 payload and its protocol history explicitly.
% \item An operational and probabilistic audit, plus a clearly separated post-hoc RF--Jev fusion analysis, yielding concrete guidance about standalone recognition, probability validation, and conditional complementary value.
% \end{itemize}

\section{Related Work}
\label{sec:related-work}

We position our work along three lines of research. We first review conventional sensor-based HAR and approaches that reduce its dependence on task-specific labeled data. We then discuss foundation-model-based, zero-shot, and training-free approaches to activity recognition. Finally, we distinguish these approaches from the general-purpose probabilistic decision model investigated in this work.

\subsection{Sensor-Based Human Activity Recognition}

Sensor-based human activity recognition has traditionally been formulated as a supervised learning problem. Measurements from wearable or mobile sensors are divided into temporal windows and used to train classifiers for a predefined set of activities. Earlier systems commonly relied on manually designed statistical and frequency-domain features, while deep learning shifted much of the representation learning into the model itself~\cite{wang2019deep,ordonez2016deep}. Despite strong recognition performance, supervised models remain dependent on the data available during training and can struggle when users, devices, sensor placements, or activity distributions change.

A substantial body of work has therefore investigated how this dependence can be reduced. Self-supervised learning, for example, uses unlabeled sensor data to learn reusable representations before adapting them to downstream HAR tasks~\cite{logacjov2024selfsupervised}. Cross-dataset approaches address a related problem by learning representations that transfer between different sensing configurations and activity spaces. GOAT combines sensor representations with natural-language supervision and evaluates this idea for cross-dataset and zero-shot activity recognition~\cite{miao2024goat}. These approaches improve the ability of HAR models to generalize, but they still learn sensor representations or adapt model parameters from sensor data.

Our work considers a different setting. We do not attempt to learn a better sensor representation or adapt a recognition model to the target dataset. Instead, we ask how much activity recognition is possible when the decision model remains fixed and only a deterministic description of the current sensor observation is provided at inference time.

\subsection{Foundation Models and Training-Free HAR}

Foundation models have recently introduced new ways of approaching sensor-based HAR. Rather than learning every recognition task from scratch, large-scale pretraining can provide representations or prior knowledge that are reused across datasets and applications. A recent survey identifies several emerging directions, including HAR-specific foundation models, adaptations of general time-series or multimodal models, and the integration of large language models into sensor-based activity recognition~\cite{bian2026foundation}. SensorLLM, for example, first aligns motion sensor signals with textual descriptions and subsequently performs task-aware tuning for HAR~\cite{li2025sensorllm}. SensorLM instead learns sensor-language representations from large-scale wearable data and supports zero-shot activity recognition among several downstream tasks~\cite{zhang2025sensorlm}. Recent benchmarking work has also started to evaluate foundation models systematically across sensor-based HAR settings. HARBench, for example, evaluates models along multiple dimensions including domain and sensor-position robustness as well as few-shot and zero-shot performance~\cite{tanigaki2026harbench}.

Zero-shot recognition has also been studied without requiring examples of every target activity during training. IMUZero combines learned sensor representations with semantic activity attributes generated by a language model to recognize unseen activities~\cite{su2025imuzero}. In a different sensing domain, ADL-LLM transforms ambient smart-home sensor events into textual descriptions and uses a general-purpose language model for zero-shot activity recognition~\cite{civitarese2025adlllm}. These approaches demonstrate that pretrained language knowledge can help bridge sensor observations and previously unseen activity concepts, although they differ in the extent to which sensor-specific models or representations are learned.

More recent work has explicitly investigated training-free inference for wearable motion data. RAG-HAR extracts statistical descriptors from sensor windows, retrieves similar labeled samples from a vector database, and provides this contextual evidence to a large language model for activity identification~\cite{sivaroopan2026raghar}. RAG-HAR+ strengthens this retrieval-first design and reduces the number of samples that require LLM inference~\cite{karunarathna2026ragharplus}. ZARA constructs a statistically grounded textual knowledge base from reference sensor data and combines retrieved evidence with an agentic LLM workflow for training-free motion time-series reasoning~\cite{li2026zara}. STELLA takes another direction by learning a compact sensor tokenizer whose outputs are projected into the embedding space of a frozen language model for activity recognition~\cite{sivaroopan2026stella}.

Together, these studies show that removing task-specific classifier training does not define a single experimental setting. A system may still learn a sensor-specific representation, use labeled reference data, retrieve examples, or rely on generative language-model reasoning. We study a complementary setting in which no sensor encoder or HAR classifier is trained for the target task and no labeled sensor examples are provided or retrieved at inference time. The fixed decision model receives only a deterministic representation of the current sensor window, a fixed task instruction, and the candidate activity set.
Evaluating pretrained general-purpose models on established HAR benchmarks introduces an additional methodological concern. Public sensor datasets may have been accessible during model pretraining, making it difficult to establish whether a benchmark is genuinely unseen. Haresamudram et al. investigated this issue using memorization tests and reported evidence that a large language model could reproduce parts of public wearable sensor datasets~\cite{haresamudram2024memorize}. We therefore do not assume that public HAR benchmarks were necessarily absent from the pretraining data of proprietary general-purpose models.

\subsection{System One and General-Purpose Decision Models}

The approaches discussed above primarily use pretrained models through learned sensor representations, retrieval mechanisms, or generative language-model inference. Activity recognition, however, ultimately requires a bounded decision among candidate activities. This raises the question of whether a general-purpose model designed directly for structured decisions can provide a different interface for training-free HAR.

TypeSafe AI introduced Jev in September 2026 as the first public model under its proposed class of System One Models. According to the manufacturer, System One Models are designed for fast, structured decisions rather than autoregressive text generation. Jev receives an input state together with predefined choices and returns typed probabilistic decisions~\cite{almeida2026jev}. Details of Jev's architecture and training data remain proprietary. We therefore treat Jev as a black-box decision model and make no assumptions about its internal operation.

Independent studies have already begun to evaluate Jev in different application domains. Rafe and Das use Jev to convert police crash narratives into probabilistic crash variables and explicitly evaluate its probabilities against coded fields and blinded human judgments~\cite{rafe2026calibrated}. Huang et al. investigate Jev as a low-cost judge of factual agreement between radiology reports~\cite{huang2026jev}. Most closely related to our use of deterministic signal-derived features, JEVQA evaluates Jev for video-quality prediction from encoding metadata, bitstream statistics, and pixel-derived features without task-specific model training~\cite{robitza2026jevqa}. These studies show that Jev can operate as a fixed decision component across substantially different input domains, while also illustrating that its uncertainty estimates require empirical evaluation rather than being assumed reliable.

Our study considers sensor-based human activity recognition. We investigate whether a fixed System One Model can map deterministic representations derived from wearable motion measurements to semantic activity classes without HAR-specific training, fine-tuning, or labeled reference examples. Unlike JEVQA, where the target is a perceptual video-quality score, our target classes describe the physical human activities that generated the measured motion signal. We separately evaluate recognition performance and the extent to which the model's reported probabilities provide useful information about prediction uncertainty.

\section{Setting and Experimental Method}
\label{sec:method}
\begin{figure*}[t]
\centering
\includegraphics[width=\textwidth]{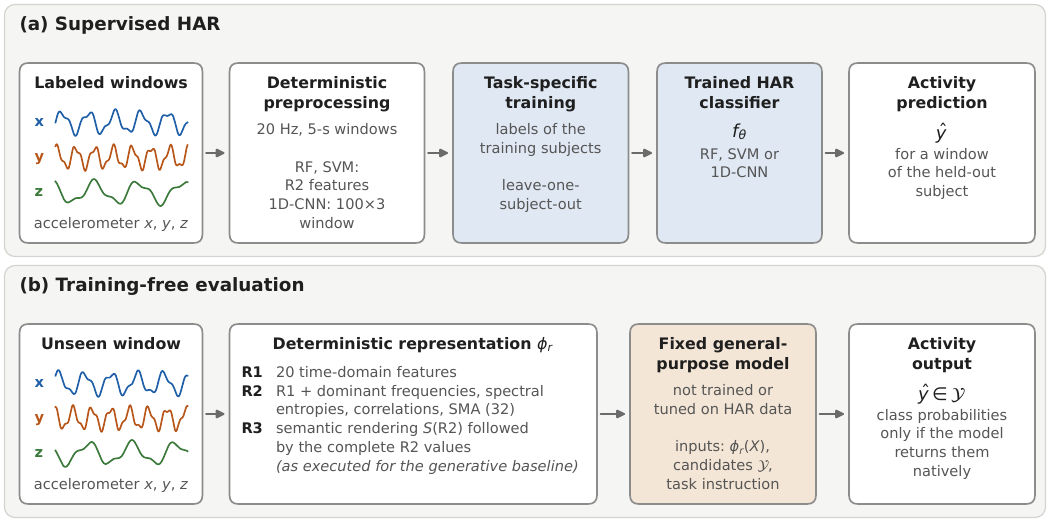}
\caption{Frozen pipeline overview, retained from the generative-baseline stage. The R3 condition shown here was subsequently adopted unchanged for Jev: semantic rendering plus complete R2 values. ``Unseen'' denotes absence from the supervised training fold or query examples; it does not establish absence from proprietary pretraining.}
\Description{The upper pipeline trains RF, SVM, or CNN on other subjects. The lower pipeline provides deterministic R1, R2, or R3 and candidate activities to a fixed general-purpose model.}
\label{fig:overview}
\end{figure*}

We design the study to separate three questions that are easily conflated in training-free HAR. First, can a fixed general-purpose
decision model recognize activities from deterministic descriptions of sensor windows at all? Second, how strongly does recognition
depend on the representation exposed to the model? Third, do the resulting decisions provide useful probabilities, operational
advantages, or complementary information beyond a trained HAR classifier? Figure~\ref{fig:overview} summarizes the supervised and
training-free evaluation paths. The remainder of this section defines the datasets, representations, models, and statistical protocol used
to answer these questions.
\subsection{Scope and Analysis Provenance}
The unit of inference is a five-second accelerometer window $X_i\in\mathbb{R}^{100\times3}$ with label $y_i\in\mathcal{Y}=\{c_1,\ldots,c_K\}$. A supervised classifier learns a decision rule from other subjects' labeled windows. Jev instead receives $\phi_r(X_i)$, the fixed task instruction $I$, and the fixed candidate vocabulary:
\begin{equation}
J_{1.13.0}\!\left(\phi_r(X_i), I, \mathcal{Y}\right)
=
\left(
\hat{y}^{\mathrm{native}}_i,
\mathbf{p}_i,
c^{\mathrm{native}}_i
\right).
\end{equation}
The native choice and probability argmax are distinct outputs in our evaluation. We do not assume that the returned vector is valid until it passes the checks below. ``Training-free'' means no HAR-specific model fitting, fine-tuning, or labeled examples in the query; it does not mean absence of provider pretraining, benchmark labels used for evaluation, or empirical preprocessing.

The main evaluation protocol was frozen in version control before external-model inference. It specifies the benchmark, metrics, bootstrap procedure, and failure rules. We refer to this as a \emph{pre-inference protocol freeze} rather than a formal preregistration because the commits were not independently timestamped in a public registry. Additional confidence and error-overlap diagnostics were specified before Jev inference but were not part of the original main protocol. The hybrid analysis was designed after the main Jev results and is therefore treated separately as \emph{post hoc and exploratory}.
% The repository calls these specifications preregistered; more precisely, they are \emph{locally committed, pre-inference protocol freezes}. Local Git ordering and hashes are not an independently timestamped public registration. Additional confidence and error-overlap diagnostics were specified before Jev outcomes but outside the original main specification. The hybrid analysis was conceived after the main Jev results and is explicitly \emph{post hoc and exploratory}, even though its complete grid and scoring rules were frozen before hybrid scoring. These three evidential categories are kept separate.

\subsection{Datasets, Preprocessing, and Benchmark Selection}
We use WISDM's smartwatch accelerometer (UCI507)~\cite{weiss2019wisdm}, the raw smartphone acceleration of UCI341/HAPT~\cite{reyesortiz2015hapt}, and PAMAP2's Protocol hand/wrist accelerometer~\cite{reiss2012pamap2,pamap2data}. UCI341 is not the conventional prewindowed UCI HAR benchmark: we use its six basic activities, exclude postural transitions, and construct new five-second windows from raw recordings. We retain the native activity taxonomies rather than mapping classes across datasets (Table~\ref{tab:datasets}).
\begin{table*}[t]
\centering
\caption{Frozen study inputs. Each window has 100 samples at 20\,Hz, spans 5\,s, and has zero overlap. ``Eligible'' is the full preprocessed population; ``benchmark'' is the matched evaluation subset.}
\label{tab:datasets}
\begin{tabular}{lrrlrrr}
\toprule
Dataset & Subjects & Classes & Selected signal & Native rate & Eligible & Benchmark\\
\midrule
WISDM (UCI507) &51&18&Watch acceleration, xyz&Timestamped&31,945&900\\
UCI341 (HAPT)&30&6&Smartphone total acceleration, xyz&50\,Hz&2,579&300\\
PAMAP2 (UCI231)&9&12&Hand/wrist $\pm16g$ acceleration, xyz&100\,Hz&3,833&600\\
\bottomrule
\end{tabular}
\end{table*}

\paragraph{Physical preprocessing.}
All channels preserve gravity and orientation; there is no rotation, body/gravity separation, or per-window normalization. WISDM exact duplicate observations are removed. Activity changes, non-increasing timestamps, or gaps of at least 500\,ms split segments. Within segments, linear interpolation produces a 20\,Hz grid. This frozen WISDM procedure has no anti-alias filter, so aliasing remains a limitation. UCI341 values are converted from $g$ using $9.80665\,\mathrm{m/s^2}$ and resampled from 50 to 20\,Hz. Its label intervals use the frozen 1-based inclusive interpretation; the source documentation does not settle endpoint indexing unambiguously. PAMAP2 uses columns 5--7 of Protocol files, already in $\mathrm{m/s^2}$, excluding code 0. Missing selected-axis triplets are linearly interpolated only when complete flanks are less than 500\,ms apart; longer gaps split segments and incomplete ends are discarded. UCI341 and PAMAP2 use fixed polyphase FIR resampling with a Kaiser window ($\beta=5$), zero boundary extension, and retained edge transients.

No window crosses an annotated activity or subject boundary. A complete window requires native observations through its five-second endpoint; its emitted grid covers 0--4.95\,s. Incomplete remainders are discarded. Activity labels thus define eligibility, segmentation, and evaluation strata, but never enter feature values, semantic threshold fitting, or model queries. This is evaluation on isolated activity segments, not continuous transition detection.

\paragraph{Matched benchmark.}
The frozen v2 benchmark contains 50 windows per class: 900 WISDM, 300 UCI341, and 600 PAMAP2 windows. Quotas are allocated round-robin across eligible subjects in ascending identifier order, then spread deterministically over each subject/class sequence. For sequence length $n$ and quota $q<n$, selected indices are $\lfloor(2j+1)(n-1)/(2q)\rfloor$, $j=0,\ldots,q-1$; an exhausted sequence contributes all windows. Thus selection is class-stratified and independent of model predictions, features, and confidence. v2 replaced an earlier earliest-window rule before any external-model inference; v1 was retained as provenance and not used for inference. Supervised predictions were joined only after benchmark selection. All recognition comparisons use the identical windows; full-population baseline results are not substituted for matched-subset scores.

\subsection{Representations: Information versus Accessibility}
\paragraph{R1: 20 numerical features.}
For each axis and the magnitude $m_t=(x_t^2+y_t^2+z_t^2)^{1/2}$, R1 contains mean, population standard deviation, minimum, maximum, and mean squared amplitude:
\begin{equation}
 \mu_x=T^{-1}\sum_t x_t,\quad
 \sigma_x=\sqrt{T^{-1}\sum_t(x_t-\mu_x)^2},\quad
 E_x=T^{-1}\sum_t x_t^2.
\end{equation}
Values are computed in float64 and serialized without feature rounding.

\paragraph{R2: 32 numerical features.}
R2 retains all R1 values unchanged and adds four dominant frequencies, four normalized spectral entropies, three pairwise Pearson correlations, and signal magnitude area. For each mean-centered channel, the untapered real DFT uses the 50 positive bins at 0.2--10\,Hz, excluding DC and including Nyquist. With $P_j=|F_j|^2$ and $q_j=P_j/\sum_{l=1}^{50}P_l$,
\begin{equation}
 f_{\rm dom}=f_{\arg\max_{1\leq j\leq50}P_j},\qquad
 H_{\rm spec}=-\frac{\sum_{j:q_j>0}q_j\ln q_j}{\ln50}.
\end{equation}
For constant signals or zero total power, frequency and entropy are zero sentinels; exact frequency ties select the lowest bin. Correlations are zero when either axis is constant. Finally,
\begin{equation}
 \mathrm{SMA}=T^{-1}\sum_t(|x_t|+|y_t|+|z_t|).
\end{equation}
The normalized entropy formula and zero-power conventions match the executed implementation.

\paragraph{R3: semantic augmentation, not semantic text alone.}
A deterministic English renderer $S$ describes all 32 features and attaches low, moderate, or high qualifiers to 20 selected quantities (weak, moderate, or strong for absolute correlations). Boundaries are the one-third and two-third quantiles of all 31,945 WISDM watch windows, fitted without activity labels and reused unchanged on UCI341 and PAMAP2. Means and extrema remain numerical. Values at a boundary enter the lower category. This is an empirical, transductive WISDM preprocessing step, not a threshold fit restricted to training subjects.

The actual input to \emph{both} general-purpose models is
\begin{equation}
 \boxed{\mathrm{R3}=S(\mathrm{R2})\oplus\mathrm{R2}},
 \label{eq:r3}
\end{equation}
where $\oplus$ denotes serialization into one input. Prose displays two decimal places (scientific notation for tiny nonzero values); the appended JSON retains complete unrounded R2 values. No activity names, definitions, or additional sensor observations are introduced by $S$.

\paragraph{Documented R3 deviation.}
An earlier representation boundary specified prose-only R3. The Claude benchmark inadvertently retained the historical audit JSON suffix. The recorded erratum verified all 1,800 Claude R3 requests against their corresponding R2 payloads; no results were replaced. Before any Jev request, Jev's protocol explicitly adopted the same complete R3 input for comparability. Accordingly, R3--R2 measures the joint effect of semantic rendering, redundancy, and serialization/length changes. It cannot establish an isolated causal effect of semantics, nor support claims about prose-only R3.

\subsection{Inference and Supervised References}
Jev uses the pinned identifier \texttt{jev-1.13.0} and the TypeSafe System One endpoint. R1/R2 are structured objects; R3 is the frozen string. One choice question supplies the native candidate names in canonical order, with no descriptions. The shared instruction asks for one activity performed during the five-second accelerometer window. No labeled examples, retrieval, external tools, or prompt tuning are used. Each dataset/window receives R1, R2, and R3, totaling 5,400 requests. Dataset/window ordering is hash-based and interleaved, with sequential requests. Transport-only retries are permitted by the protocol; HTTP-200 answers are never replaced because of content or correctness. The completed Jev run required no retries.

Claude Sonnet 5 is the generative baseline, queried through the Anthropic API with structured output restricted to the same activity vocabulary, the same instruction, windows, and representation payloads. Its saved outputs provide labels, not native class-probability vectors suitable for this analysis. We therefore do not invent Claude calibration estimates or compare verbal confidence to Jev probabilities.

Supervised baselines use leave-one-subject-out (LOSO) evaluation on the complete eligible population. RF uses 500 trees, square-root feature subsampling, balanced class weights, and seed 2026 on R2. SVM uses train-fold standardization, an RBF kernel, $C=1$, scale-based gamma, and balanced class weights, without probability calibration. The CNN consumes the raw $100\times3$ windows with train-fold axis standardization: convolution blocks of 64/128/128 channels (kernels 5/5/3), batch normalization and ReLU, two max-pooling layers, global average pooling, and a linear output. Adam uses learning rate $10^{-3}$, weight decay $10^{-4}$, batch size 128, and inverse-frequency class weights. Training stops within 100 epochs with patience 10; a cyclically selected separate validation subject determines the checkpoint, never the test subject. The initial CNN smoke stage was followed by the completed frozen full-LOSO evaluation. All reported CNN scores use that completed evaluation. No retuning was performed for the matched subset or hybrid analysis.

\subsection{Recognition, Probabilities, and Statistical Uncertainty}
\label{sec:metrics}
Recognition uses Jev's native choice on every benchmark window. Missing, failed, refused, or invalid choices would count as incorrect and reduce true-class recall; none occurred. Primary macro-F1 averages across the entire native taxonomy with undefined class F1 set to zero; accuracy is secondary:
\begin{equation}
 F_{\rm macro}=K^{-1}\sum_k F1_k,\qquad
 \mathrm{Acc}=N^{-1}\sum_i\mathbb{1}[\hat y_i^{\rm native}=y_i].
\end{equation}

A Jev probability vector must cover each candidate exactly once, be finite and in $[0,1]$, and satisfy $|\sum_kp_{ik}-1|\leq10^{-6}$. Invalid vectors are excluded \emph{only} from probabilistic metrics and counted; they are never clamped, imputed, or renormalized. A valid native choice remains in recognition scoring even if its probability vector is invalid. Probabilistic analyses use $\hat y_i^p=\arg\max_k p_{ik}$ with canonical-order tie breaking and $c_i=\max_kp_{ik}$, rather than the provider's separate native confidence.

On the valid-vector set $V$, we report multiclass Brier score~\cite{brier1950}, clipped negative log-likelihood, and ten-bin expected calibration error (ECE)~\cite{guo2017calibration}:
\begin{align}
 \mathrm{BS}&=|V|^{-1}\sum_{i\in V}\sum_k(p_{ik}-\mathbb{1}[y_i=c_k])^2,\\
 \mathrm{NLL}&=-|V|^{-1}\sum_{i\in V}\ln\max(p_{i,y_i},10^{-12}),\\
 \mathrm{ECE}&=\sum_b\frac{|B_b|}{|V|}|\mathrm{acc}(B_b)-\overline c(B_b)|.
\end{align}
Bins are $[0,0.1),\ldots,[0.9,1]$, with empty-bin counts retained. Clipping is confined to NLL evaluation and does not repair a vector. Selective prediction~\cite{geifman2017selective} uses every unique observed confidence threshold:
\begin{align}
 A_\tau&=\{i\in V:c_i\geq\tau\},\quad \mathrm{coverage}(\tau)=|A_\tau|/|V|,\\
 \mathrm{risk}(\tau)&=|A_\tau|^{-1}\sum_{i\in A_\tau}\mathbb{1}[\hat y_i^p\ne y_i].
\end{align}
Coverage is conditional on valid vectors, not all API requests. The area under the stepwise risk--coverage curve (AURC) is descriptive. No confidence threshold is fitted for deployment.

Accuracy and macro-F1 intervals use a subject-cluster bootstrap. We use 10,000 replicates and seed 2026 within each dataset. Each replicate draws the original number of subjects with replacement and includes all their benchmark windows with multiplicity. We compute pooled metrics, not mean subject scores; absent classes stay in the macro-F1 denominator. Percentile bounds use linear interpolation. Paired comparisons share draws and windows. Intervals are descriptive, conditional on the fixed benchmark and observed model responses; they are neither simultaneous confidence bands nor evidence from repeated API runs. No significance tests or claims of superiority adjusted for multiple comparisons are made.

\subsection{Operational Measurements and Research Assistance}
Latency is observed request-to-response wall time, including network/service effects, at concurrency one. It excludes sensor acquisition and preprocessing. Costs use recorded tokens and the list prices frozen at execution, not invoices: Jev USD 0.042 per million input tokens with output free; Claude USD 2 per million input and USD 10 per million output tokens. We report the frozen median and 95th percentile, plus Jev's observed maximum; the available scored summaries do not contain a 99th percentile. No new timing experiment is performed.

Generative AI tools (ChatGPT/Codex and Claude) were used for language revision, implementation and code checking, analysis
support, and the preparation of manuscript tables and figures from recorded results. The research question, experimental design,
scientific decisions, interpretation, and responsibility for the work remained with the author. The evaluated Jev and Claude outputs are
experimental observations and are distinct from the use of these tools for research assistance. All numerical findings reported in this
paper derive from the frozen executable analyses; AI-generated suggestions were not treated as empirical evidence.

\section{Main Results}
\label{sec:results}

We organize the main results around three aspects of the training-free evaluation. We first compare recognition performance with the generative and supervised references. We then examine how the sensor representation changes Jev's predictions and evaluate whether its returned probabilities provide useful uncertainty information. Finally, we report the observed latency and API cost. The post-hoc fusion analysis is kept separate in Section 5 because it was designed after the main Jev results were known.

\subsection{Standalone Recognition: A Large Supervised Gap}
Table~\ref{tab:main} reports all main recognition conditions with their frozen intervals. Jev R3 achieves macro-F1 0.0378 on WISDM, 0.1184 on UCI341, and 0.0893 on PAMAP2. The corresponding RF values are 0.7057, 0.8899, and 0.8041. Thus the Jev--RF deficits are 0.6679, 0.7715, and 0.7148 macro-F1, respectively. Even the strongest Jev condition remains far below all three supervised references. These results do not support using Jev as a drop-in replacement for a trained HAR classifier under the evaluated conditions.

\begin{figure*}[t]
\centering
\includegraphics[width=\textwidth]{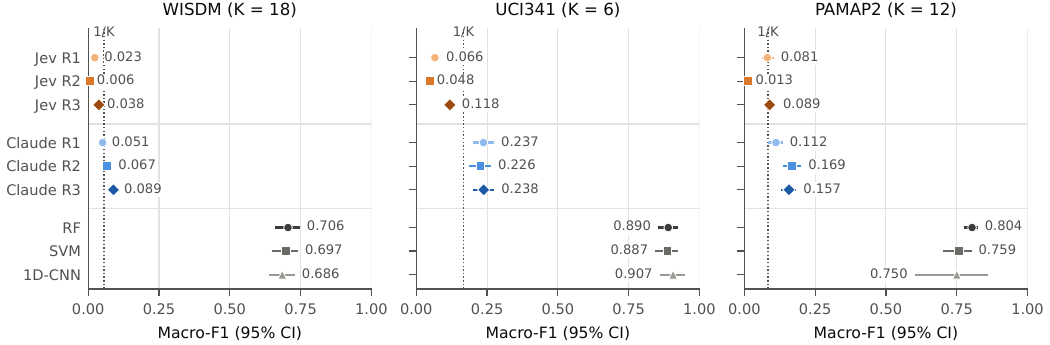}
\caption{Matched-benchmark macro-F1 for Jev, the Claude generative baseline, and supervised HAR references, with 95\%
subject-cluster bootstrap intervals. Both training-free models use the same frozen benchmark windows and representations R1--R3; supervised values are obtained from frozen LOSO predictions on the same windows. The dotted line marks $1/K$ as a scale reference. Jev's point estimates remain below Claude's across all nine dataset--representation pairs and far below the supervised references. Exact values are reported in Table~\ref{tab:main}.}
\Description{Three dataset panels compare Claude under R1, R2, and R3 with Random Forest, SVM, and CNN; supervised recognition is substantially higher.}
\label{fig:recognition}
\end{figure*}

Claude also remains far below supervised recognition, although its point estimate exceeds Jev in every matched combination of representation and dataset. The paired PAMAP2 R1 difference between Jev and Claude has a 95\% interval spanning zero; we do not claim universal statistical superiority. The matched comparison of Jev, Claude, and the supervised references in Figure~\ref{fig:recognition} provides context for the weakness of both training-free interfaces. Published retrieval-based and sensor-pretrained systems use materially different information budgets and protocols, so we do not rank them numerically against these results.

\begin{table*}[t]
\centering\small
\caption{All matched main recognition results, reported as estimate [95\% subject-cluster interval]. Scores are proportions. Jev uses native choice on every window; the other systems use their frozen class predictions.}
\label{tab:main}
\begin{tabular}{lll ll}
\toprule Dataset & Model & Representation & Accuracy [95\% CI] & Macro-F1 [95\% CI]\\\midrule
WISDM & Jev & R1 & 0.0733 [0.0653, 0.0822] & 0.0231 [0.0153, 0.0320]\\
WISDM & Claude & R1 & 0.1278 [0.1173, 0.1386] & 0.0507 [0.0428, 0.0587]\\
WISDM & Jev & R2 & 0.0556 [0.0546, 0.0562] & 0.0059 [0.0058, 0.0059]\\
WISDM & Claude & R2 & 0.1422 [0.1288, 0.1569] & 0.0666 [0.0536, 0.0815]\\
WISDM & Jev & R3 & 0.1122 [0.1001, 0.1237] & 0.0378 [0.0341, 0.0414]\\
WISDM & Claude & R3 & 0.1467 [0.1321, 0.1616] & 0.0891 [0.0757, 0.1019]\\
WISDM & RF & native & 0.7056 [0.6567, 0.7489] & 0.7057 [0.6589, 0.7475]\\
WISDM & SVM & native & 0.7000 [0.6488, 0.7455] & 0.6970 [0.6483, 0.7407]\\
WISDM & 1D-CNN & native & 0.6856 [0.6355, 0.7317] & 0.6857 [0.6377, 0.7300]\\
\midrule
UCI341 & Jev & R1 & 0.1767 [0.1667, 0.1889] & 0.0657 [0.0537, 0.0799]\\
UCI341 & Claude & R1 & 0.3300 [0.2943, 0.3673] & 0.2372 [0.2012, 0.2749]\\
UCI341 & Jev & R2 & 0.1667 [0.1667, 0.1667] & 0.0476 [0.0476, 0.0476]\\
UCI341 & Claude & R2 & 0.2933 [0.2569, 0.3298] & 0.2263 [0.1859, 0.2641]\\
UCI341 & Jev & R3 & 0.2367 [0.2118, 0.2619] & 0.1184 [0.0989, 0.1363]\\
UCI341 & Claude & R3 & 0.3267 [0.2847, 0.3693] & 0.2379 [0.2011, 0.2751]\\
UCI341 & RF & native & 0.8900 [0.8542, 0.9252] & 0.8899 [0.8536, 0.9250]\\
UCI341 & SVM & native & 0.8867 [0.8435, 0.9252] & 0.8869 [0.8438, 0.9253]\\
UCI341 & 1D-CNN & native & 0.9067 [0.8587, 0.9487] & 0.9067 [0.8591, 0.9483]\\
\midrule
PAMAP2 & Jev & R1 & 0.1567 [0.1386, 0.1807] & 0.0812 [0.0631, 0.1038]\\
PAMAP2 & Claude & R1 & 0.1800 [0.1369, 0.2356] & 0.1116 [0.0836, 0.1356]\\
PAMAP2 & Jev & R2 & 0.0833 [0.0754, 0.0939] & 0.0129 [0.0118, 0.0145]\\
PAMAP2 & Claude & R2 & 0.2367 [0.1931, 0.2949] & 0.1686 [0.1374, 0.2008]\\
PAMAP2 & Jev & R3 & 0.1617 [0.1426, 0.1983] & 0.0893 [0.0711, 0.1116]\\
PAMAP2 & Claude & R3 & 0.2433 [0.2036, 0.2977] & 0.1575 [0.1289, 0.1835]\\
PAMAP2 & RF & native & 0.8033 [0.7763, 0.8250] & 0.8041 [0.7761, 0.8241]\\
PAMAP2 & SVM & native & 0.7550 [0.7055, 0.7959] & 0.7586 [0.7025, 0.8021]\\
PAMAP2 & 1D-CNN & native & 0.7433 [0.5763, 0.8555] & 0.7504 [0.6014, 0.8607]\\
\bottomrule\end{tabular}\end{table*}

\subsection{Representation: Numerical Enrichment Can Fail}
R2 contains all R1 features, yet Jev macro-F1 decreases on every dataset. R2 assigns walking to 899/900 WISDM windows, 300/300 UCI341 windows, and 595/600 PAMAP2 windows. With exactly 50 examples per class, single-class prediction has accuracy $1/K$ and macro-F1 $2/[K(K+1)]$; the UCI341 R2 scores are exactly this collapsed solution. Its zero-width bootstrap interval reflects constant behavior under the benchmark's subject/class composition, not precise evidence of useful generalization.

R3 reverses part of this loss. Paired R3--R2 macro-F1 differences are 0.0319 [0.0283, 0.0355] on WISDM, 0.0708 [0.0513, 0.0887] on UCI341, and 0.0763 [0.0579, 0.0987] on PAMAP2. Relative to R1, gains are 0.0147 [0.0050, 0.0236], 0.0527 [0.0333, 0.0718], and 0.0080 [$-0.0072$, 0.0219], respectively. The PAMAP2 R3--R1 interval therefore does not support a clear improvement. R3 adds no new measurement information, but changes how existing information is presented. Its partial recovery supports the practical importance of the interface while leaving the mechanism unresolved.

\subsection{Probability Validity Does Not Imply Calibration}
All 5,400 Jev choices are valid, but 86 probability vectors fail the frozen criterion, leaving 5,314 for probability evaluation (Table~\ref{tab:prob}). We preserve this distinction: a successful typed answer is not proof of a usable distribution. Across valid vectors, native choice and canonical probability argmax disagree in 96 cases, including 22, 2, and 29 R3 cases on WISDM, UCI341, and PAMAP2. The hybrid endpoint must consequently be compared with probability-argmax recognition, not the main native-choice score.

R3 ECE is 0.2532, 0.3489, and 0.1294, respectively. Mean maximum probabilities are 0.3667, 0.5844, and 0.2952, while argmax accuracies on valid vectors are only 0.1135, 0.2405, and 0.1658. Seven WISDM R3 cases require the frozen NLL floor; no other condition does. Confidence cannot be read as a validated probability of correctness. Smaller ECE on PAMAP2 also does not establish a generally better model: ECE depends on confidence distribution, binning, and task composition.

\begin{table*}[t]\centering\small
\caption{Frozen Jev probability metrics on valid vectors only; no repair. ``Native $\neq$ argmax'' counts disagreements between the
native choice and the canonical probability argmax. BS, NLL, ECE, and AURC are point estimates; the frozen analysis does not provide
intervals for these metrics.}
\label{tab:prob}
\begin{tabular}{llrrrrrrr}\toprule Dataset & Rep. & Valid/total & Native $\neq$ argmax & BS & NLL & ECE & AURC & argmax acc.\\\midrule
WISDM & R1 & 882/900 & 17 & 1.0282 & 3.3020 & 0.2495 & 0.8408 & 0.0737\\
WISDM & R2 & 883/900 & 0 & 1.1350 & 3.4987 & 0.4154 & 0.8908 & 0.0555\\
WISDM & R3 & 890/900 & 22 & 1.0571 & 3.5944 & 0.2532 & 0.8600 & 0.1135\\
UCI341 & R1 & 297/300 & 4 & 0.9523 & 1.9385 & 0.3665 & 0.7022 & 0.1751\\
UCI341 & R2 & 292/300 & 0 & 1.1045 & 2.2255 & 0.5018 & 0.7334 & 0.1610\\
UCI341 & R3 & 291/300 & 2 & 0.9400 & 1.8631 & 0.3489 & 0.6306 & 0.2405\\
PAMAP2 & R1 & 590/600 & 21 & 0.9274 & 2.5515 & 0.1043 & 0.8223 & 0.1593\\
PAMAP2 & R2 & 598/600 & 1 & 1.0108 & 2.7647 & 0.2903 & 0.8845 & 0.0836\\
PAMAP2 & R3 & 591/600 & 29 & 0.9319 & 2.5462 & 0.1294 & 0.8012 & 0.1658\\
\bottomrule\end{tabular}\end{table*}
\begin{figure*}[t]
\centering
\includegraphics[width=\textwidth]{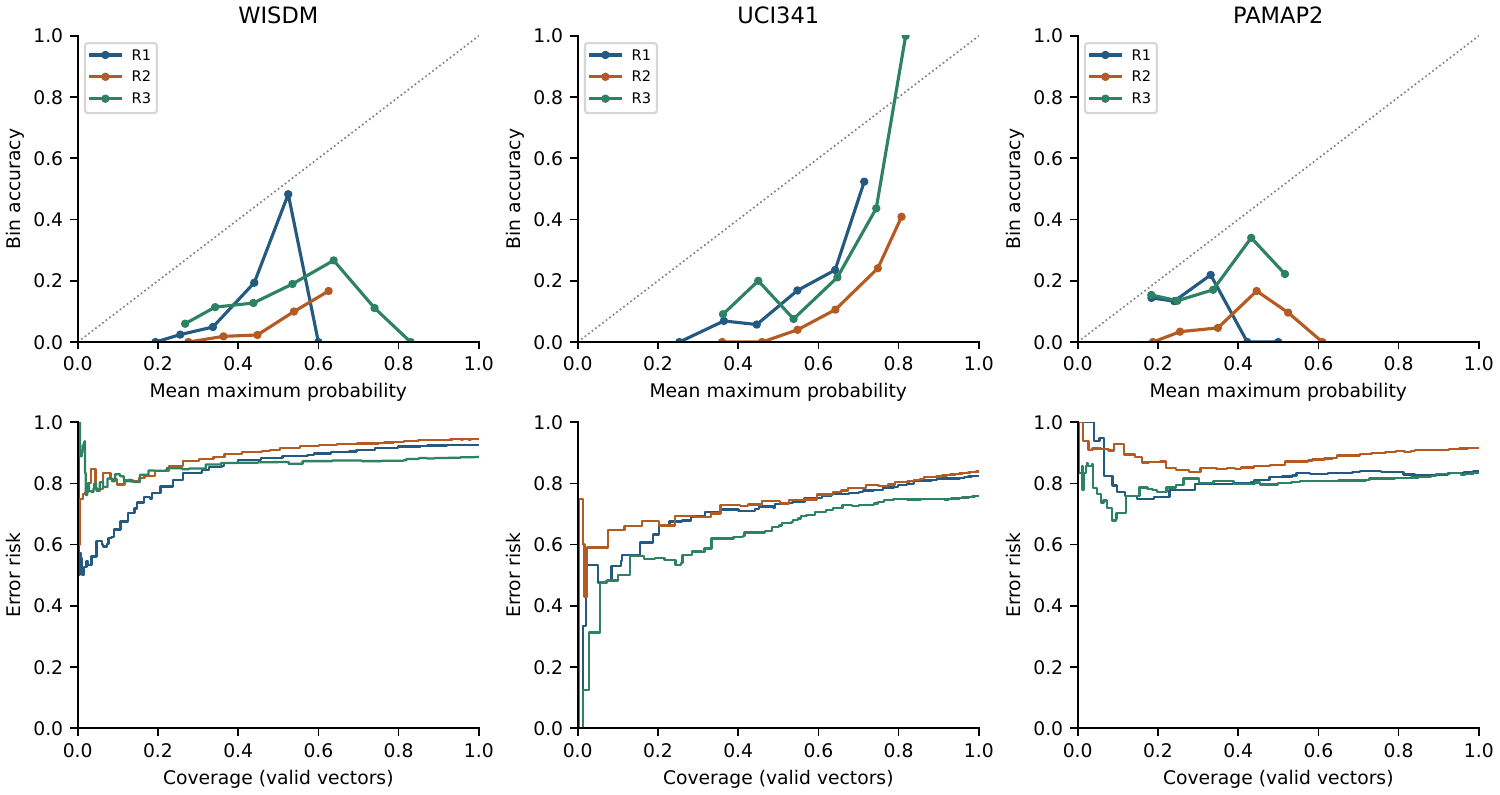}
\caption{Jev probability audit, rendered directly from frozen reliability bins and risk--coverage points. Upper panels show empirical accuracy versus mean maximum probability in nonempty bins; the diagonal denotes perfect calibration. Lower panels show risk conditional on a valid vector, at every frozen confidence threshold. No threshold is selected and no confidence calibration is fitted.}
\Description{Three columns for WISDM, UCI341, and PAMAP2, with reliability diagrams above risk coverage curves. R1, R2, and R3 have substantial errors despite nonzero confidence.}
\label{fig:probability}
\end{figure*}

Selective rejection can change the error rate, but the full curves in Figure~\ref{fig:probability} give no general operating guarantee. AURC for R3 is 0.8600, 0.6306, and 0.8012. On WISDM, R1 has a lower AURC (0.8408) than R3 despite lower full-coverage recognition. Representation quality for classification and confidence ranking therefore need not improve together. In the additional prospective diagnostics, R3 maximum-confidence AUROC for argmax correctness is 0.5803, 0.6750, and 0.5859. These are diagnostics outside the original main specification, not fitted acceptance policies.

\subsection{Cost and Latency: Feasible Requests, Weak Decisions}
\begin{table*}[t]\centering\small
\caption{Observed sequential API latency and frozen list-price cost. Each row is a dataset/representation condition. Costs are USD per 1,000 windows, calculated from logged usage; timing excludes acquisition and preprocessing.}
\label{tab:ops}
\begin{tabular}{llrrrrrr}\toprule
&&\multicolumn{4}{c}{Jev}&\multicolumn{2}{c}{Claude}\\
Dataset & Rep. & Median (s)&p95 (s)&Max (s)&USD/1k&Median (s)&USD/1k\\\midrule
WISDM & R1 & 0.296 & 0.668 & 5.327 & 0.0376 & 1.457 & 1.6342\\
WISDM & R2 & 0.292 & 0.561 & 2.225 & 0.0492 & 1.471 & 2.0587\\
WISDM & R3 & 0.300 & 0.561 & 1.930 & 0.0707 & 1.519 & 3.5469\\
UCI341 & R1 & 0.292 & 0.480 & 1.184 & 0.0348 & 1.489 & 1.5446\\
UCI341 & R2 & 0.292 & 0.507 & 5.335 & 0.0461 & 1.491 & 1.9632\\
UCI341 & R3 & 0.293 & 0.516 & 1.704 & 0.0677 & 1.558 & 3.4533\\
PAMAP2 & R1 & 0.293 & 0.516 & 2.001 & 0.0359 & 1.440 & 1.5607\\
PAMAP2 & R2 & 0.291 & 0.579 & 1.670 & 0.0474 & 1.448 & 1.9859\\
PAMAP2 & R3 & 0.295 & 0.622 & 1.559 & 0.0690 & 1.531 & 3.4745\\
\bottomrule\end{tabular}\end{table*}
Jev's nine condition-level medians span 0.291--0.300\,s, compared with 1.440--1.558\,s for Claude. Jev costs USD 0.035--0.071 per 1,000 windows at the recorded list price, versus USD 1.545--3.547 for Claude. The complete 5,400-case Jev run costs USD 0.2777 on that basis; the Claude run costs USD 12.8168. These are observed run-specific API comparisons, not controlled hardware benchmarks or guaranteed future prices.

All Jev requests completed without retries or invalid choices. Nonetheless, the slowest response was 5.335\,s, exceeding the five-second window duration. A subsecond median is compatible with periodic querying, but does not demonstrate deadline guarantees, streaming throughput under load, on-device inference, offline operation, or low energy consumption. Window accumulation already contributes five seconds before a decision can be requested. The supervised models' local inference and training costs have different boundaries; we make no API-cost claim of superiority over RF deployment.

\section{Post-Hoc Exploratory Probability Fusion}
\label{sec:hybrid}
This analysis asks whether weak standalone Jev evidence can complement a specialized classifier. Both its conception and the choice of R3 were informed by the completed main results. It is not part of the pre-inference main evaluation protocol. Its protocol, full weight grid, arithmetic, exclusions, and bootstrap were frozen before hybrid scoring; there was no further Jev inference or retuning.

For the same window, the linear probability pool is
\begin{align}
 \mathbf p_{H,i}(\lambda)&=(1-\lambda)\mathbf p_{RF,i}+\lambda\mathbf p_{Jev,i},\label{eq:hybrid}\\
 \hat y_{H,i}(\lambda)&=\arg\max_{c_k\in\mathcal Y}p_{H,i,k}(\lambda),\\
 \Lambda&=\{k/20:k=0,\ldots,20\}.
\end{align}
Arithmetic is float64; ties select the first maximum in canonical class order. This is a linear ensemble of two outputs conditioned on the observation. The informal phrase ``Jev as a prior'' must not be mistaken for Bayesian prior--likelihood updating or calibrated posterior inference.

Previously frozen RF models were re-executed solely to export probabilities. The gate required exact reproduction of all 38,357 full-LOSO hard predictions across 90 folds and all 1,800 benchmark predictions, including fold hashes and equality of RF predict and probability argmax. The saved verification reports zero mismatches. The fusion uses the saved probability export. RF vectors must be finite, in range, complete, and sum to one within $10^{-9}$; Jev uses its original $10^{-6}$ criterion.

One fixed complete-case subset per dataset is used for RF alone, Jev probability argmax alone, and \emph{all} 21 weights: 890/900 WISDM, 291/300 UCI341, and 591/600 PAMAP2 windows. The excluded 10, 9, and 9 cases have invalid Jev R3 vectors; none are repaired. All subjects remain represented. The endpoints reproduce RF and Jev argmax on these exact subsets. They need not equal Table~\ref{tab:main}, which uses all cases and native Jev choice. Bootstrap draws are paired across every weight within a dataset.

\begin{figure*}[t]
\centering
\includegraphics[width=\textwidth]{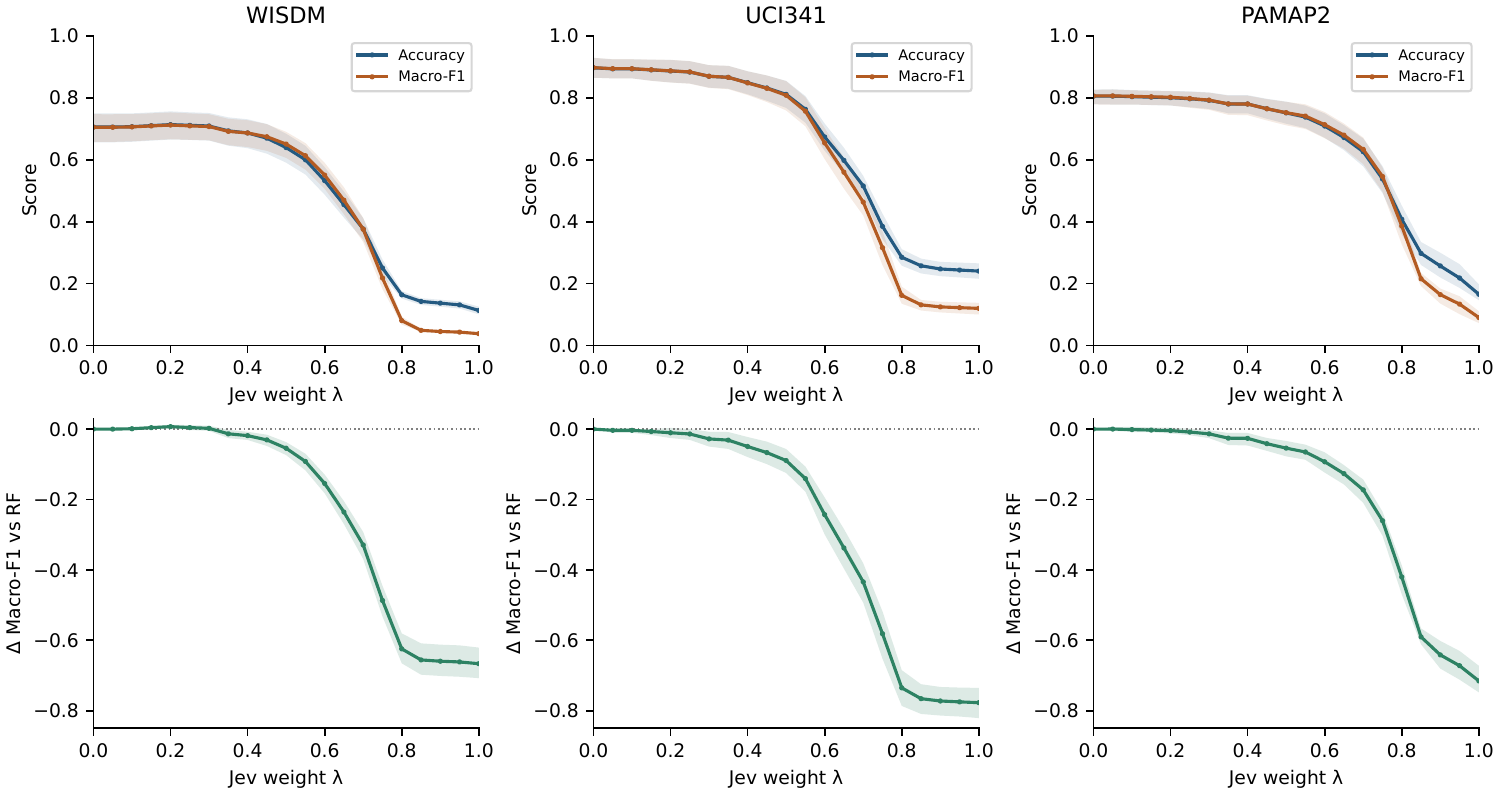}
\caption{Post-hoc exploratory fusion: complete frozen 21-weight curves on the same complete-case windows for each dataset. Shading shows pointwise 95\% subject-cluster intervals, not simultaneous confidence bands. No optimal weight is fitted or selected. The second row shows paired macro-F1 differences from RF for all weights; increasing Jev weight eventually causes large losses.}
\Description{Three columns show accuracy and macro F1 fusion curves and paired macro F1 differences. A small positive region appears on WISDM; UCI341 and PAMAP2 do not show a comparable replicated benefit.}
\label{fig:hybrid}
\end{figure*}

The full sensitivity curve is the result (Figure~\ref{fig:hybrid}; all points and intervals in Appendix~\ref{app:hybrid}). On WISDM, the RF endpoint has macro-F1 0.7050. At the descriptive example $\lambda=0.20$, macro-F1 is 0.7120 and accuracy is 0.7135. The paired gains are 0.0070 [0.0024, 0.0125] macro-F1 and 0.0079 [0.0033, 0.0136] accuracy. At this weight, eight RF errors become correct and one correct prediction becomes an error, a net seven correct decisions among 890 windows. This narrow improvement is practically much smaller than the standalone gap.

The same $\lambda=0.20$ produces macro-F1 changes of $-0.0103$ [$-0.0253$, 0.0034] on UCI341 and $-0.0043$ [$-0.0125$, 0.0042] on PAMAP2. PAMAP2's tiny positive macro-F1 change at $\lambda=0.05$ is 0.00017 [$-0.00454$, 0.00525], with unchanged accuracy; it is not a replication of the WISDM result. Larger Jev weights substantially degrade all three datasets. We neither hide unfavorable grid points nor define a fitted $\lambda^\star$. A favorable pointwise interval discovered on this exploratory curve does not validate a deployment weight or establish benefit on new subjects or datasets.

\section{Discussion}
\label{sec:discussion}

Taken together, the results point to a distinction between information contained in a sensor representation and information that a fixed general-purpose model can use. The large supervised gap shows that the evaluated representations retain substantial activity information, while the behavior across R1, R2, and R3 shows that access to this information depends strongly on the interface presented to the model. We discuss this implication first, then consider possible reasons for the differences across datasets and the practical consequences for future HAR systems.

\subsection{The Sensor Interface Is Part of the Model Evaluation}
The study's clearest practical finding is the gap between possessing sensor descriptors and using them effectively. RF receives the same R2 features that lead Jev toward near-single-class prediction. Thus these descriptors contain useful predictive structure for a trained mapping, while the tested fixed decision interface fails to exploit much of it. R3 changes how the same measurement-derived information is presented to the model and partially recovers recognition. This shows that the sensor-to-model interface should be treated as part of the experimental design.

Several mechanisms could explain this behavior, including numerical processing, associations with feature names, serialization format, ordering, redundancy, and input length. R3 changes several of these factors at once, and we did not evaluate prose-only, length-matched, or reordered variants. We therefore cannot attribute the recovery specifically to semantic language. What the experiment does show is that changing the representation materially changes model behavior even when the underlying sensor measurements remain unchanged.

% It does not demonstrate a particular internal failure mechanism. Numeric processing limitations, associations with feature names, the text/object distinction, ordering, redundancy, or input length could contribute. R3 bundles several changes and there is no prose-only, length-matched, or reordered-input ablation. Provider descriptions cannot substitute for these missing controls. The defensible claim is that representation changes materially alter behavior under the frozen interface, not that semantic language has been causally isolated as the remedy.

\subsection{Why Datasets Differ: Interpretations to Test}
UCI341's six basic activities and smartphone placement differ from WISDM's eighteen activities at the wrist. It is plausible that coarse posture and locomotion distinctions are easier to associate with summary acceleration features than fine-grained hand-related activities, whose identity may depend on context or object interaction absent from the signal. This may help explain Jev's higher UCI341 R3 macro-F1. However, class count, placement, taxonomy, subject composition, recording conditions, and preprocessing all vary together. Raw cross-dataset scores cannot isolate any of them, and class count changes trivial accuracy as well.

PAMAP2 has twelve activities and only nine subjects, with unequal activity coverage. Some activities may have distinctive intensity or periodicity, while others may share wrist motion patterns. These are plausible reasons for heterogeneous recognition, not measured causal effects. Its small number of subject clusters also limits interval stability. The semantic thresholds were developed from WISDM's full unlabeled feature population and transferred unchanged. Device/placement and distribution differences could change what ``low'' or ``high'' communicates on the other datasets. We did not refit thresholds to test this explanation.

The WISDM fusion gain could reflect occasional complementary probability rankings where RF margins are small. On WISDM, even a weak standalone classifier can therefore contribute probability rankings that alter a small number of RF decisions beneficially. UCI341 starts from a stronger RF endpoint and may offer fewer useful corrections; PAMAP2 has a different error structure and population. These are hypotheses. The observed correction/error counts support a local trade-off, not a proven mechanism or a rule that WISDM-like data will necessarily benefit. Dataset name, number of classes, and sensor placement alone are insufficient to predict success. Several WISDM candidate strings are terse (e.g., ``teeth'', ``soup'', and ``catch'') rather than full activity descriptions; their interpretation is another uncontrolled interface difference.

\subsection{What HAR Researchers Can Use and Should Avoid}
For a system with labeled training data and a recognition-quality requirement, these results favor a task-trained sensor classifier. Jev offers a fast, inexpensive typed API and requires no task-specific fitting, but those operational advantages do not compensate for the observed standalone error rates. The evidence does not justify using it as the sole activity recognizer, a trusted pseudo-label generator, or a safety-critical decision component in this setting.

Exploratory augmentation should therefore retain a supervised reference and validate every returned probability vector. Native decisions and probability argmax should be evaluated separately, and fusion endpoints should be compared on identical complete cases. Any mixture weight or rejection threshold would require independent validation before operational use. Our results provide no support for transferring the WISDM weight directly to another dataset.

% Exploratory augmentation requires independent validation of complementary evidence. Keep a supervised reference, validate every probability vector, distinguish native decisions from probability argmax, and compare fusion endpoints on identical complete cases. A mixture weight and any rejection threshold would need independent validation before operational use. Copying the WISDM example weight into another dataset is specifically unsupported by our results.

For interface design, start by auditing prediction distributions and errors, not by assuming more features must help. Preserve physical units and representation provenance. Disclose any use of unlabeled target distributions, labels for segmentation, and retrieved examples so that ``training-free'' does not hide an information advantage. For real-time systems, include acquisition delay, network tails, service availability, and fallback behavior; median API latency alone does not establish a real-time recognizer.

\section{Limitations, Ethics, and Reproducibility}
\label{sec:limitations}
\paragraph{Scope and inference.}
We evaluate one pinned Jev version, one generative baseline, three public datasets, three frozen interfaces, and one completed response per case. We do not assess alternative prompts, orderings, sensors, languages, personalized models, or version-to-version variability. Proprietary pretraining data are unknown; public benchmark contamination cannot be excluded~\cite{haresamudram2024memorize}. The results constrain these tested pipelines, not every future general-purpose decision model. No trial of other configurations is implied by our interpretations.

\paragraph{Sampling and deployment.}
Class balancing makes rare and common activities equally represented, unlike everyday prevalence. Labels define clean activity boundaries and exclude short transitions. Five-second windows and summary features discard temporal detail. WISDM threshold fitting includes the unlabeled evaluation population, so it is transductive rather than a fully inductive pipeline. WISDM interpolation and the external datasets' FIR resampling differ; edge effects and possible WISDM aliasing remain. UCI341 lacks timestamps for detecting hidden acquisition gaps. Complete-case probability analyses may be selection-biased and describe only valid responses. Subject-cluster intervals do not cover model-service variability or all sources of experimental uncertainty. Exploratory fusion and pointwise intervals cannot establish a general prior benefit.

\paragraph{Ethical scope.}
This study reuses existing public sensor datasets; it recruits no new participants and collects no new human measurements. General-purpose APIs receive derived sensor representations and activity names, without subject identifiers or ground-truth labels. Derived motion features may still contain personal information; omission of identifiers does not establish anonymity. The study does not attempt identification or clinical decision making. Original dataset documentation and access terms govern reuse. No new ethics approval was obtained for this methodological secondary analysis of publicly available data; no formal institutional exemption is claimed. The author reports no external funding or supporting project and
declares no conflicts of interest. TypeSafe AI did not fund or support this study, and access to Jev was obtained through the publicly available API and paid for by the author.

\paragraph{Reproducibility and provenance.}
The prepared artifact separates frozen preprocessing, benchmark selection, inference requests/responses, main scoring, additional diagnostics, and post-hoc fusion. Recorded model identifiers, exact serialized requests, candidate order, dependency versions, window keys, checksums, and validation manifests support traceability. The manuscript's tables and new plots are rendered from already-scored CSV/JSON artifacts; no model query, training, bootstrap rerun, or tuning is needed for this manuscript build. Existing vector figures are preserved. Appendix~\ref{app:artifact} maps claims to artifact paths; the prepared artifact index contains detailed commit and checksum records outside the main narrative. Code, evaluation scripts, and derived artifacts supporting the reported results have been prepared for public release. The corresponding archival record will be made available separately.

\section{Conclusion}
Jev is fast and inexpensive to query, but it does not replace trained HAR classifiers in the evaluated setting. Adding numerical signal features can reduce recognition, while augmenting the same numerical evidence with a deterministic semantic rendering can partially recover it. This shows that the sensor-to-model interface must be evaluated rather than treated as a neutral preprocessing step. The small exploratory fusion gain on WISDM further shows that weak standalone recognition does not necessarily imply that a model contains no complementary information. However, the gain does not replicate on UCI341 or PAMAP2 and therefore does not support a general fusion strategy.
More broadly, our results suggest that the sensor interface is part of the model. Providing more signal descriptors does not guarantee that a general-purpose decision model can turn them into more usable evidence. Fast and inexpensive inference is valuable, but it does not remove the need to validate how physical sensor measurements are represented and interpreted.

\bibliographystyle{ACM-Reference-Format}
\bibliography{references}
\clearpage
\onecolumn
\appendix
\section{Complete Exploratory Fusion Results}
\label{app:hybrid}
The following tables reproduce every frozen weight, point estimate, and 95\% interval. All scores are proportions; each dataset has one fixed complete-case subset. Differences are paired against that subset's RF endpoint. These are post-hoc exploratory, pointwise intervals, not multiplicity-adjusted evidence or a selected model.
\begin{table}[htbp]\centering\small
\caption{WISDM: all exploratory fusion weights. Scores and paired differences are proportions with 95\% subject-cluster intervals.}
\begin{tabular}{rllll}\toprule $\lambda$ & Accuracy & Macro-F1 & $\Delta$ Accuracy & $\Delta$ Macro-F1\\\midrule
0.00 & 0.7056 [0.6562, 0.7494] & 0.7050 [0.6578, 0.7469] & 0.0000 [0.0000, 0.0000] & 0.0000 [0.0000, 0.0000]\\
0.05 & 0.7056 [0.6562, 0.7494] & 0.7049 [0.6577, 0.7469] & 0.0000 [0.0000, 0.0000] & -0.0001 [-0.0004, 0.0000]\\
0.10 & 0.7067 [0.6581, 0.7497] & 0.7060 [0.6596, 0.7474] & 0.0011 [0.0000, 0.0034] & 0.0010 [-0.0003, 0.0034]\\
0.15 & 0.7101 [0.6618, 0.7528] & 0.7091 [0.6632, 0.7498] & 0.0045 [0.0011, 0.0091] & 0.0041 [0.0006, 0.0087]\\
0.20 & 0.7135 [0.6652, 0.7561] & 0.7120 [0.6662, 0.7527] & 0.0079 [0.0033, 0.0136] & 0.0070 [0.0024, 0.0125]\\
0.25 & 0.7112 [0.6640, 0.7531] & 0.7093 [0.6644, 0.7496] & 0.0056 [-0.0011, 0.0127] & 0.0043 [-0.0024, 0.0113]\\
0.30 & 0.7090 [0.6618, 0.7508] & 0.7072 [0.6628, 0.7470] & 0.0034 [-0.0034, 0.0104] & 0.0022 [-0.0050, 0.0095]\\
0.35 & 0.6933 [0.6455, 0.7368] & 0.6917 [0.6470, 0.7322] & -0.0124 [-0.0236, -0.0022] & -0.0133 [-0.0256, -0.0021]\\
0.40 & 0.6865 [0.6374, 0.7306] & 0.6864 [0.6411, 0.7275] & -0.0191 [-0.0302, -0.0089] & -0.0185 [-0.0308, -0.0075]\\
0.45 & 0.6697 [0.6196, 0.7154] & 0.6743 [0.6288, 0.7149] & -0.0360 [-0.0517, -0.0213] & -0.0307 [-0.0479, -0.0159]\\
0.50 & 0.6393 [0.5906, 0.6840] & 0.6506 [0.6060, 0.6901] & -0.0663 [-0.0858, -0.0480] & -0.0543 [-0.0739, -0.0369]\\
0.55 & 0.6000 [0.5522, 0.6451] & 0.6131 [0.5681, 0.6537] & -0.1056 [-0.1297, -0.0828] & -0.0919 [-0.1172, -0.0699]\\
0.60 & 0.5326 [0.4872, 0.5761] & 0.5504 [0.5061, 0.5900] & -0.1730 [-0.1996, -0.1472] & -0.1546 [-0.1826, -0.1292]\\
0.65 & 0.4551 [0.4130, 0.4972] & 0.4693 [0.4238, 0.5108] & -0.2506 [-0.2819, -0.2190] & -0.2357 [-0.2700, -0.2047]\\
0.70 & 0.3764 [0.3422, 0.4112] & 0.3758 [0.3345, 0.4121] & -0.3292 [-0.3641, -0.2926] & -0.3292 [-0.3692, -0.2924]\\
0.75 & 0.2506 [0.2248, 0.2778] & 0.2181 [0.1826, 0.2512] & -0.4551 [-0.4936, -0.4141] & -0.4869 [-0.5325, -0.4451]\\
0.80 & 0.1640 [0.1519, 0.1761] & 0.0806 [0.0678, 0.0934] & -0.5416 [-0.5817, -0.4960] & -0.6243 [-0.6654, -0.5803]\\
0.85 & 0.1427 [0.1315, 0.1531] & 0.0493 [0.0447, 0.0545] & -0.5629 [-0.6051, -0.5147] & -0.6557 [-0.6977, -0.6086]\\
0.90 & 0.1371 [0.1257, 0.1481] & 0.0456 [0.0422, 0.0490] & -0.5685 [-0.6109, -0.5205] & -0.6594 [-0.7012, -0.6123]\\
0.95 & 0.1315 [0.1196, 0.1430] & 0.0435 [0.0399, 0.0472] & -0.5742 [-0.6170, -0.5255] & -0.6614 [-0.7035, -0.6143]\\
1.00 & 0.1135 [0.1010, 0.1257] & 0.0385 [0.0347, 0.0424] & -0.5921 [-0.6336, -0.5461] & -0.6665 [-0.7075, -0.6207]\\
\bottomrule\end{tabular}\end{table}
\begin{table}[htbp]\centering\small
\caption{UCI341: all exploratory fusion weights. Scores and paired differences are proportions with 95\% subject-cluster intervals.}
\begin{tabular}{rllll}\toprule $\lambda$ & Accuracy & Macro-F1 & $\Delta$ Accuracy & $\Delta$ Macro-F1\\\midrule
0.00 & 0.8969 [0.8648, 0.9288] & 0.8974 [0.8648, 0.9291] & 0.0000 [0.0000, 0.0000] & 0.0000 [0.0000, 0.0000]\\
0.05 & 0.8935 [0.8627, 0.9241] & 0.8940 [0.8627, 0.9246] & -0.0034 [-0.0108, 0.0000] & -0.0034 [-0.0106, 0.0000]\\
0.10 & 0.8935 [0.8627, 0.9241] & 0.8940 [0.8627, 0.9246] & -0.0034 [-0.0108, 0.0000] & -0.0034 [-0.0106, 0.0000]\\
0.15 & 0.8900 [0.8552, 0.9231] & 0.8905 [0.8547, 0.9234] & -0.0069 [-0.0169, 0.0000] & -0.0069 [-0.0170, 0.0000]\\
0.20 & 0.8866 [0.8497, 0.9221] & 0.8870 [0.8492, 0.9226] & -0.0103 [-0.0254, 0.0035] & -0.0103 [-0.0253, 0.0034]\\
0.25 & 0.8832 [0.8462, 0.9181] & 0.8835 [0.8455, 0.9185] & -0.0137 [-0.0304, 0.0000] & -0.0139 [-0.0306, 0.0001]\\
0.30 & 0.8694 [0.8328, 0.9052] & 0.8696 [0.8317, 0.9054] & -0.0275 [-0.0488, -0.0070] & -0.0278 [-0.0496, -0.0075]\\
0.35 & 0.8660 [0.8294, 0.9030] & 0.8660 [0.8279, 0.9032] & -0.0309 [-0.0552, -0.0076] & -0.0314 [-0.0563, -0.0083]\\
0.40 & 0.8488 [0.8114, 0.8857] & 0.8480 [0.8089, 0.8855] & -0.0481 [-0.0766, -0.0217] & -0.0494 [-0.0794, -0.0225]\\
0.45 & 0.8316 [0.7905, 0.8720] & 0.8306 [0.7864, 0.8721] & -0.0653 [-0.0959, -0.0350] & -0.0668 [-0.0995, -0.0356]\\
0.50 & 0.8110 [0.7675, 0.8547] & 0.8081 [0.7591, 0.8537] & -0.0859 [-0.1159, -0.0554] & -0.0893 [-0.1247, -0.0565]\\
0.55 & 0.7629 [0.7196, 0.8057] & 0.7567 [0.7080, 0.8018] & -0.1340 [-0.1661, -0.1031] & -0.1406 [-0.1779, -0.1065]\\
0.60 & 0.6735 [0.6291, 0.7167] & 0.6543 [0.6020, 0.7017] & -0.2234 [-0.2702, -0.1794] & -0.2431 [-0.2993, -0.1942]\\
0.65 & 0.5979 [0.5548, 0.6386] & 0.5597 [0.5066, 0.6066] & -0.2990 [-0.3495, -0.2525] & -0.3377 [-0.3965, -0.2838]\\
0.70 & 0.5155 [0.4798, 0.5482] & 0.4634 [0.4203, 0.4998] & -0.3814 [-0.4339, -0.3333] & -0.4340 [-0.4928, -0.3811]\\
0.75 & 0.3849 [0.3432, 0.4262] & 0.3161 [0.2595, 0.3687] & -0.5120 [-0.5700, -0.4585] & -0.5813 [-0.6525, -0.5178]\\
0.80 & 0.2852 [0.2581, 0.3114] & 0.1623 [0.1352, 0.1903] & -0.6117 [-0.6632, -0.5629] & -0.7351 [-0.7864, -0.6848]\\
0.85 & 0.2577 [0.2331, 0.2809] & 0.1316 [0.1134, 0.1475] & -0.6392 [-0.6877, -0.5931] & -0.7658 [-0.8092, -0.7247]\\
0.90 & 0.2474 [0.2243, 0.2704] & 0.1250 [0.1072, 0.1415] & -0.6495 [-0.6944, -0.6060] & -0.7724 [-0.8137, -0.7325]\\
0.95 & 0.2440 [0.2201, 0.2680] & 0.1225 [0.1037, 0.1398] & -0.6529 [-0.6979, -0.6083] & -0.7749 [-0.8165, -0.7343]\\
1.00 & 0.2405 [0.2158, 0.2656] & 0.1203 [0.1005, 0.1381] & -0.6564 [-0.7036, -0.6105] & -0.7771 [-0.8210, -0.7352]\\
\bottomrule\end{tabular}\end{table}
\begin{table}[htbp]\centering\small
\caption{PAMAP2: all exploratory fusion weights. Scores and paired differences are proportions with 95\% subject-cluster intervals.}
\begin{tabular}{rllll}\toprule $\lambda$ & Accuracy & Macro-F1 & $\Delta$ Accuracy & $\Delta$ Macro-F1\\\midrule
0.00 & 0.8054 [0.7795, 0.8265] & 0.8061 [0.7786, 0.8255] & 0.0000 [0.0000, 0.0000] & 0.0000 [0.0000, 0.0000]\\
0.05 & 0.8054 [0.7782, 0.8276] & 0.8063 [0.7776, 0.8266] & 0.0000 [-0.0046, 0.0050] & 0.0002 [-0.0045, 0.0052]\\
0.10 & 0.8037 [0.7783, 0.8248] & 0.8048 [0.7783, 0.8241] & -0.0017 [-0.0081, 0.0063] & -0.0013 [-0.0084, 0.0067]\\
0.15 & 0.8020 [0.7764, 0.8234] & 0.8033 [0.7768, 0.8225] & -0.0034 [-0.0086, 0.0034] & -0.0028 [-0.0089, 0.0046]\\
0.20 & 0.8003 [0.7750, 0.8222] & 0.8018 [0.7755, 0.8211] & -0.0051 [-0.0128, 0.0029] & -0.0043 [-0.0125, 0.0042]\\
0.25 & 0.7970 [0.7696, 0.8208] & 0.7978 [0.7679, 0.8195] & -0.0085 [-0.0167, 0.0000] & -0.0083 [-0.0174, 0.0011]\\
0.30 & 0.7919 [0.7640, 0.8173] & 0.7926 [0.7613, 0.8165] & -0.0135 [-0.0246, -0.0032] & -0.0135 [-0.0256, -0.0032]\\
0.35 & 0.7800 [0.7500, 0.8084] & 0.7803 [0.7450, 0.8070] & -0.0254 [-0.0423, -0.0094] & -0.0257 [-0.0451, -0.0109]\\
0.40 & 0.7800 [0.7500, 0.8084] & 0.7799 [0.7438, 0.8069] & -0.0254 [-0.0423, -0.0094] & -0.0262 [-0.0466, -0.0110]\\
0.45 & 0.7648 [0.7324, 0.7967] & 0.7646 [0.7269, 0.7955] & -0.0406 [-0.0584, -0.0232] & -0.0415 [-0.0618, -0.0252]\\
0.50 & 0.7513 [0.7165, 0.7877] & 0.7520 [0.7112, 0.7870] & -0.0541 [-0.0748, -0.0333] & -0.0540 [-0.0774, -0.0337]\\
0.55 & 0.7377 [0.7020, 0.7748] & 0.7411 [0.6994, 0.7788] & -0.0677 [-0.0851, -0.0483] & -0.0650 [-0.0872, -0.0442]\\
0.60 & 0.7090 [0.6708, 0.7495] & 0.7133 [0.6695, 0.7540] & -0.0964 [-0.1202, -0.0725] & -0.0928 [-0.1234, -0.0659]\\
0.65 & 0.6717 [0.6309, 0.7116] & 0.6795 [0.6357, 0.7166] & -0.1337 [-0.1623, -0.1045] & -0.1265 [-0.1573, -0.1023]\\
0.70 & 0.6261 [0.5786, 0.6685] & 0.6335 [0.5862, 0.6716] & -0.1794 [-0.2166, -0.1439] & -0.1725 [-0.2095, -0.1433]\\
0.75 & 0.5381 [0.4925, 0.5793] & 0.5459 [0.4968, 0.5748] & -0.2673 [-0.3066, -0.2252] & -0.2601 [-0.3029, -0.2344]\\
0.80 & 0.4078 [0.3586, 0.4521] & 0.3860 [0.3268, 0.4206] & -0.3976 [-0.4395, -0.3492] & -0.4201 [-0.4696, -0.3869]\\
0.85 & 0.2978 [0.2604, 0.3345] & 0.2158 [0.1922, 0.2347] & -0.5076 [-0.5424, -0.4592] & -0.5903 [-0.6115, -0.5662]\\
0.90 & 0.2572 [0.2199, 0.2998] & 0.1646 [0.1359, 0.1842] & -0.5482 [-0.5940, -0.4900] & -0.6415 [-0.6804, -0.6017]\\
0.95 & 0.2183 [0.1863, 0.2629] & 0.1339 [0.1015, 0.1593] & -0.5871 [-0.6282, -0.5289] & -0.6722 [-0.7118, -0.6302]\\
1.00 & 0.1658 [0.1465, 0.1979] & 0.0907 [0.0727, 0.1110] & -0.6396 [-0.6724, -0.5857] & -0.7154 [-0.7479, -0.6723]\\
\bottomrule\end{tabular}\end{table}

\clearpage
\section{Artifact Map and Audit Boundaries}
\label{app:artifact}
\begin{table}[htbp]
\centering\small
\caption{Compact artifact map. Paths are relative to the study repository; full hashes and provenance reside in the accompanying artifact index and frozen manifests.}
\begin{tabularx}{\textwidth}{lX}
\toprule
Evidence & Artifact or directory\\\midrule
Preprocessing/representations & \path{docs/protocol_v1.md}; \path{docs/dataset_adaptation_protocol_v1.md}; \path{configs/semantic_thresholds_v1.yaml}\\
Benchmark and deviations & \path{docs/training_free_evaluation_protocol_v2.md}; \path{docs/training_free_v2_r3_protocol_deviation.md}\\
Jev request contract & \path{docs/jev_inference_protocol_v1.md}\\
Main scores and intervals & \path{results/training_free_v2/scored_jev/summary.csv}; \path{paired_differences.csv} in the same directory\\
Probability audit & \path{probabilistic.csv}; \path{probabilistic_details.json} in the main scoring directory\\
Operational evidence & \path{operational.csv} in the main scoring directory and \path{results/training_free_v2/scored/}\\
Exploratory fusion & \path{docs/hybrid_protocol_v1.md}; \path{results/exploratory/hybrid_v1/scored/}\\
Independent frozen verification & \path{results/training_free_v2/scored_jev_validation_v1.json}; \path{results/exploratory/hybrid_v1/scored_validation_v1.json}\\
\bottomrule
\end{tabularx}
\end{table}
The main probability audit contains all reliability-bin counts and threshold-level coverage/risk values. Full paired comparisons, class metrics, and confusion matrices remain in the frozen score artifacts. The lightweight source package supplies the material needed to compile this manuscript; it does not redistribute the original human sensor archives or claim to contain provider model weights.
\subsection{Frozen Candidate Vocabulary}
The following strings were supplied without definitions, in the listed order. Spelling and case are part of the frozen interface.
\paragraph{WISDM.} \texttt{walking}, \texttt{jogging}, \texttt{stairs}, \texttt{sitting}, \texttt{standing}, \texttt{typing}, \texttt{teeth}, \texttt{soup}, \texttt{chips}, \texttt{pasta}, \texttt{drinking}, \texttt{sandwich}, \texttt{kicking}, \texttt{catch}, \texttt{dribbling}, \texttt{writing}, \texttt{clapping}, \texttt{folding}.
\paragraph{UCI341.} \texttt{WALKING}, \texttt{WALKING\_UPSTAIRS}, \texttt{WALKING\_DOWNSTAIRS}, \texttt{SITTING}, \texttt{STANDING}, \texttt{LAYING}.
\paragraph{PAMAP2.} \texttt{lying}, \texttt{sitting}, \texttt{standing}, \texttt{walking}, \texttt{running}, \texttt{cycling}, \texttt{Nordic walking}, \texttt{ascending stairs}, \texttt{descending stairs}, \texttt{vacuum cleaning}, \texttt{ironing}, \texttt{rope jumping}.
\end{document}